\documentclass{fairmeta} 

\usepackage[utf8]{inputenc}
\usepackage{url}
\usepackage{verbatim}
\usepackage{xspace}
\usepackage{enumitem}
\usepackage{textcomp}
\usepackage{comment}
\usepackage{titlesec}
\usepackage{hyperref}

\usepackage{graphicx}
\usepackage{wrapfig}
\usepackage{adjustbox}
\usepackage{booktabs}
\usepackage{multirow}
\usepackage{array}
\usepackage{colortbl}
\usepackage{tabularx}

\newcolumntype{x}[1]{>{\centering\arraybackslash}p{#1pt}}
\newcolumntype{y}[1]{>{\raggedright\arraybackslash}p{#1pt}}
\newcolumntype{z}[1]{>{\raggedleft\arraybackslash}p{#1pt}}

\newlength\savewidth

\usepackage{amsmath, amssymb}
\usepackage{bm} 
\usepackage{pifont}
\usepackage{bbding}

\usepackage{xcolor}

\definecolor{adptorange}{RGB}{248, 205, 172}
\definecolor{cmpblue}{RGB}{189, 215, 238}
\definecolor{our_red}{RGB}{232,157,160}
\definecolor{our_blue}{RGB}{136,206,230}
\definecolor{our_orange}{RGB}{246,200,168}
\definecolor{our_green}{RGB}{178,211,164}
\definecolor{attn_code0}{RGB}{247,215,200}
\definecolor{attn_code1}{RGB}{238,169,139}
\definecolor{mlp_code0}{RGB}{204,201,221}
\definecolor{mlp_code1}{RGB}{102,95,153}
\definecolor{token_blue}{RGB}{84, 120, 140}
\definecolor{codeblue}{rgb}{0.25, 0.5, 0.5}
\definecolor{codekw}{rgb}{0.35, 0.35, 0.75}
\definecolor{green}{HTML}{009000}
\definecolor{red}{HTML}{ea4335}
\definecolor{lightgray}{gray}{0.95}
\definecolor{darkblue}{rgb}{0.1,0.1,0.6}
\definecolor{darkgreen}{rgb}{0.0,0.5,0.0}
\definecolor{darkred}{rgb}{0.6,0.1,0.1}
\definecolor{paleblue}{RGB}{250, 253, 255}
\definecolor{palegreen}{RGB}{250, 255, 250}
\definecolor{palecream}{RGB}{255, 253, 250}

\usepackage{algorithm}
\usepackage{algorithmic}
\usepackage{listings}
\usepackage{caption}
\usepackage{float} 

\lstdefinestyle{Pytorch}{
    language = Python,
    backgroundcolor = \color{white},
    basicstyle = \fontsize{9pt}{8pt}\selectfont\ttfamily\bfseries,
    columns = fullflexible,
    aboveskip=1pt,
    belowskip=1pt,
    breaklines = true,
    captionpos = b,
    commentstyle = \color{codeblue},
    keywordstyle = \color{codekw},
}

\renewcommand{\paragraph}[1]{\vspace{1.25mm}\noindent\textbf{#1}}

\makeatletter
\DeclareRobustCommand\onedot{\futurelet\@let@token\@onedot}
\def\@onedot{\ifx\@let@token.\else.\null\fi\xspace}

\makeatother

\title{UniLumos: Fast and Unified Image and Video Relighting with Physics-Plausible Feedback}

\author{%
  Ropeway Liu$^{1,2,*}$,
  Hangjie Yuan$^{\dagger2,3,1,*}$,
  Bo Dong$^{2,3}$, 
  Jiazheng Xing$^{1,2,4}$,
  Jinwang Wang$^{2,3,1}$,
  Rui Zhao$^{4}$,\\
  \textbf{Yan Xing}$^{2,3}$\textbf{,} 
  \textbf{Weihua Chen}$^{\dagger2,3}$\textbf{,} 
  \textbf{Fan Wang}$^{2}$\\
  $^1$Zhejiang University,  
  $^2$DAMO Academy, Alibaba Group,
  $^3$Hupan Lab,
  $^4$National University of Singapore\\
  $*$ Equal contributions.
  $^\dagger$ Corresponding author.\\
  \texttt{\{yuanhangjie.yhj, kugang.cwh\}@alibaba-inc.com}
}

\abstract{
Relighting is a crucial task with both practical demand and artistic value, and recent diffusion models have shown strong potential by enabling rich and controllable lighting effects. 
However, as they are typically optimized in semantic latent space, where proximity does not guarantee physical correctness in visual space, they often produce unrealistic results—such as overexposed highlights, misaligned shadows, and incorrect occlusions. 
We address this with \textbf{UniLumos}, a unified relighting framework for both images and videos that brings RGB-space geometry feedback into a flow-matching backbone. 
By supervising the model with depth and normal maps extracted from its outputs, we explicitly align lighting effects with the scene structure, enhancing physical plausibility.
Nevertheless, this feedback requires high-quality outputs for supervision in visual space, making standard multi-step denoising computationally expensive. To mitigate this, we employ path consistency learning, allowing supervision to remain effective even under few-step training regimes. 
To enable fine-grained relighting control and supervision, we design a structured six-dimensional annotation protocol capturing core illumination attributes.  
Building upon this, we propose \textbf{LumosBench}, a disentangled attribute-level benchmark that evaluates lighting controllability via large vision-language models, enabling automatic and interpretable assessment of relighting precision across individual dimensions.
Extensive experiments demonstrate that UniLumos achieves state-of-the-art relighting quality with significantly improved physical consistency, while delivering a 20x speedup for both image and video relighting.
Code is available at \href{https://github.com/alibaba-damo-academy/Lumos-Custom}{https://github.com/alibaba-damo-academy/Lumos-Custom}.
}
\date{\today}

\begin{document}
\maketitle

\begin{figure*}[t]
\centering
\includegraphics[width=0.9\textwidth]{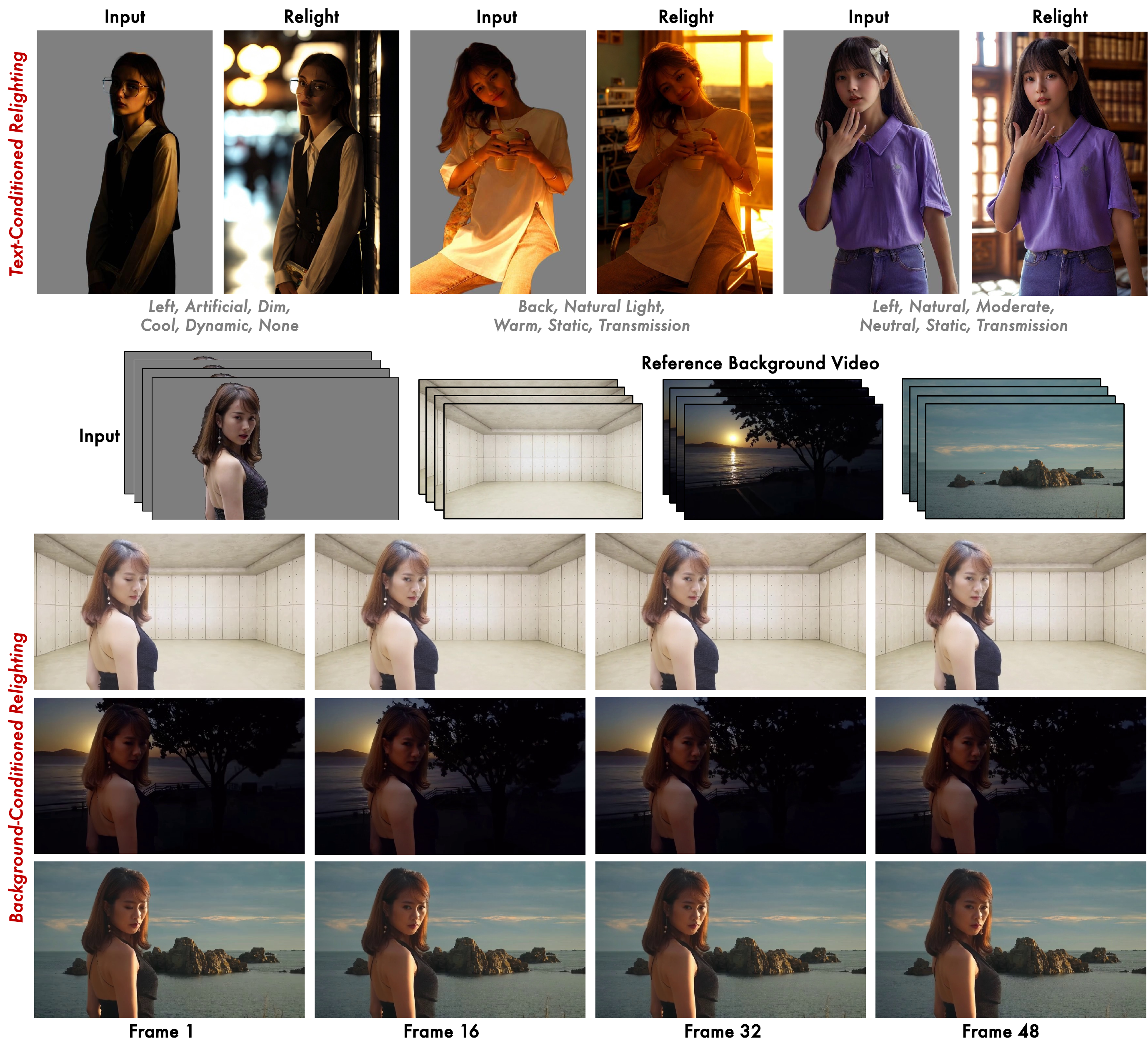}
\vspace{-2mm}
\caption{UniLumos performs physically plausible image and video relighting, conditioned on textual prompts and reference videos.}
\label{effect:fig_0}
\vspace{-4mm}
\end{figure*}

\section{Introudction}
Relighting, altering illumination in images or videos while preserving intrinsic scene attributes such as geometry, reflectance, and content, is a longstanding problem in computer vision and graphics \cite{ren2015image, zhou2019deep}. 
It underpins a wide range of applications in film production, gaming, and augmented reality, where seamless lighting integration is critical to visual fidelity.
Beyond realism, lighting conveys rich aesthetic and semantic cues—it defines atmosphere, evokes emotion, and reinforces narrative structure.
Relighting is thus not only a technical challenge but also a creative tool that shapes how characters, objects, and environments are perceived. 
However, achieving physically consistent lighting remains a core challenge—requiring the alignment of illumination effects with different scene attributes across both space and time.
To address this, traditional approaches often rely on inverse rendering pipelines~\cite{zhang2021physg, zhang2022modeling, blattmann2023stable} that estimate intrinsic scene properties, such as geometry, reflectance, and environmental lighting, from input images. 
While these methods provide physically grounded results, they typically require complex inputs—such as high dynamic range images or spherical harmonics coefficients—and are limited to constrained domains. 
This makes them impractical for real-world applications, where users often provide only a single image, a short video, or a high-level lighting prompt as input.
These limitations underscore the need for a new paradigm—one that can deliver high-quality, physically plausible relighting while operating under minimal and naturalistic input conditions.

Recent diffusion-based relighting methods~\cite{zhang2025scaling, chaturvedi2025synthlight, fang2025relightvid, zhou2025light} have shown promise by leveraging large-scale image and video datasets to produce diverse and controllable lighting effects under various user-defined conditions, such as reference images or text prompts. 
However, this strength reveals a fundamental weakness: diffusion models typically operate in semantic latent space, where similarity does not guarantee physical correctness in the visual domain.
As a result, they often fail to respect scene geometry, especially in complex scenes with dynamic lighting or temporal constraints.
For example, IC-Light~\cite{zhang2025scaling} and SynthLight~\cite{chaturvedi2025synthlight}, which are primarily designed for image relighting, lack both temporal modeling and explicit physical supervision in the visual domain.
Instead, they rely on latent-space representations, such as MLP-based embeddings (IC-Light) or multi-stage training (SynthLight), which often lead to artifacts like misaligned shadows, overexposed highlights, or incorrect lighting directions—particularly under complex geometry or extreme illumination. 
Light-A-Video~\cite{fang2025relightvid} and RelightVid~\cite{fang2025relightvid} extend these methods to the video domain, aiming to improve temporal coherence while retaining the visual quality of diffusion-based relighting. 
Light-A-Video is a training-free framework that combines IC-Light with a pre-trained video diffusion model via iterative alignment. While this improves frame-wise consistency, it incurs high inference costs due to multiple model passes. RelightVid adopts a joint training strategy with a video diffusion backbone, which enhances temporal stability compared to training-free approaches. 
However, it still operates without explicit physical supervision, resulting in inaccurate light-scene interactions and limited generalization to complex or dynamic environments. 
In short, existing diffusion-based methods excel in synthesizing plausible appearances but fall short in enforcing the physical plausibility that is essential for high-quality relighting.

To bridge the gap between generative flexibility and physical correctness, we propose \textbf{UniLumos}, a unified relighting framework for both images and videos that brings RGB-space geometry feedback into a flow-matching backbone. 
Unlike existing diffusion-based approaches that operate purely in latent space, UniLumos introduces a physics-plausible feedback mechanism that supervises generation with dense geometric signals—specifically, depth and surface normals—estimated from its outputs. 
These lighting-invariant cues serve as ideal supervision signals, enabling the model to explicitly align illumination with the scene structure, significantly improving shadow alignment, shading consistency, and spatial coherence.
Nevertheless, this feedback requires high-quality outputs for supervision in visual space, making standard multi-step denoising computationally expensive. To mitigate this, we employ path consistency learning~\cite{frans2024one}, allowing supervision to remain effective even under few-step training regimes. 

Beyond model-level improvements, existing relighting methods lack structured illumination descriptions and dedicated evaluation metrics. Generic generation scores (e.g., FID, LPIPS) fail to capture lighting-specific errors such as shadow misalignment, intensity mismatch, or incorrect light direction. 
To address this, we introduce LumosData, a scalable data pipeline that extracts diverse relighting pairs from real-world videos. 
At its core is a structured six-dimensional annotation protocol covering direction, light source type, intensity, color temperature, temporal dynamics, and optical phenomena—enabling both fine-grained conditioning during training and physically grounded evaluation at test time.
Building upon this, we propose \textbf{LumosBench}, a disentangled attribute-level benchmark that evaluates lighting controllability via large vision-language models, enabling automatic and interpretable assessment of relighting precision across individual dimensions.

Our contributions are summarized as follows: 
\begin{itemize} 
    \item \textbf{Unified Relighting with Physics-Plausible Feedback:} We propose UniLumos, a unified relighting framework for both images and videos that incorporates RGB-space geometry feedback into a flow-matching backbone, explicitly aligning lighting effects with the scene structure to enhance the physical plausibility of relighting. 
    \item \textbf{Structured Illumination Annotation and Evaluation Benchmark:} We design a structured six-dimensional annotation protocol that captures core illumination attributes, enabling fine-grained control and supervision. Building upon this, we introduce LumosBench, a disentangled attribute-level benchmark that leverages large vision-language models to automatically and interpretably evaluate relighting controllability across individual lighting dimensions.    
    \item \textbf{Extensive Validation:} Extensive experiments demonstrate that UniLumos achieves state-of-the-art relighting quality with significantly improved physical consistency, while delivering a 20x speedup for both image and video relighting. 
\end{itemize}

\section{Related Work}
\noindent\textbf{Video Diffusion Models.} Recent advances in video diffusion models~\cite{SVD_2023_arxiv, LatentDiffusion_2023_cvpr, Videocrafter1_2023_arxiv, wan2025} have enabled the generation of temporally coherent video sequences conditioned on various inputs such as text or images. In the field of text-to-video (T2V) generation~\cite{show1_2024_ijcv}, most methods extend existing text-to-image diffusion backbones with additional modules that capture temporal dynamics across frames. In contrast, a few approaches train video diffusion models from scratch to directly learn spatiotemporal priors~\cite{wan2025}. 
For image-to-video (I2V) tasks, where static images are animated with plausible motion, several methods propose specialized architectures tailored for image animation \cite{siarohin2019first, xu2024magicanimate}. 
Other strategies offer lightweight, plug-and-play adapters that can be integrated into pre-trained models. Stable Video Diffusion \cite{blattmann2023stable}, for example, fine-tunes T2V models for I2V tasks, achieving state-of-the-art performance. Beyond synthesis quality, a growing body of work emphasizes controllability, allowing users to guide generation with fine-grained constraints \cite{zhao2023uni, zhang2023adding}. 

\noindent\textbf{Relighting Methods.} Recent advances in deep neural networks have significantly improved lighting control for 2D and 3D visual content, especially in portrait relighting. Methods such as Relightful Harmonization \cite{ren2024relightful}, SwitchLight \cite{kim2024switchlight}, ConceptSliders \cite{gandikota2024concept}, Intrinsic Image Diffusion \cite{kocsis2024intrinsic}, Neural Gaffer \cite{jin2024neural}, DI-Light \cite{zeng2024dilightnet}, SynthLight~\cite{chaturvedi2025synthlight}, and IC-Light \cite{zhang2025scaling} demonstrate progress in realism and controllability. While numerous portrait relighting approaches exist \cite{pandey2021total, zhang2021neural, yeh2022learning, cai2024real}, most of them rely heavily on portrait-specific priors. In contrast, UniLumos is designed as a general-purpose relighting framework that is not constrained to any particular object category. With the rise of diffusion-based generative models, approaches like LumiSculpt \cite{zhang2024lumisculpt} extend lighting control to text-to-video (T2V) generation. Moreover, RelightVid~\cite{fang2025relightvid} and Light-A-Video~\cite{zhou2025light} implemented video relighting based on IC-Light. However, achieving both precise lighting control and high visual quality in video remains challenging due to the trade-off between spatial realism and temporal consistency. 

\noindent\textbf{Feedback Learning in Generative Models.}
Feedback learning has become a powerful tool to improve output alignment in generative models, from language systems trained with human preferences~\cite{stiennon2020learning, ouyang2022training} to visual diffusion models guided by aesthetic or attribute-based rewards~\cite{chen2024id,yuan2024instructvideo,wada2024polos,lee2023aligningT2I}, \textit{e.g.}, InstructVideo~\cite{yuan2024instructvideo} and DRaFT~\cite{clark2023DRaFT}.
In visual domains, feedback can also be physical—for example, using geometric cues to guide generation toward realism. Recent advances in distillation and consistency training~\cite{gou2021knowledge, salimans2022progressive, liu2022flow, wang2023videolcm, zheng2024trajectory, frans2024one} have accelerated diffusion inference by reducing the number of denoising steps, enabling models to recover high-quality RGB outputs with just a few iterations.
However, most existing techniques focus on appearance synthesis and overlook geometry-aware feedback, which typically requires high-fidelity outputs and is incompatible with few-step inference. UniLumos bridges this gap by combining physically grounded supervision with path-consistency learning, enabling efficient and physically plausible relighting under fast sampling regimes.

\section{Preliminaries}
\noindent\textbf{Problem Formulation.}  
Given an image or video $\mathbf{S}_1 \in \mathbb{R}^{T \times H \times W \times C}$ with intrinsic scene properties (e.g., geometry, reflectance, content) under initial illumination $\mathbf{L}_1$, the goal of relighting is to modify the illumination within a subject region specified by a binary mask $\mathbf{M} \in \{0,1\}^{T \times H \times W}$ to match a target lighting condition $\mathbf{C}$.  
The condition $\mathbf{C}$ may take the form of an image, video, or text description and implicitly defines a desired illumination field $\mathbf{L}_2$.  
The relit output $\mathbf{S}_2 \in \mathbb{R}^{T \times H \times W \times C}$ should exhibit lighting consistent with $\mathbf{L}_2$ in the masked region $\mathbf{M}$ while preserving the intrinsic attributes of $\mathbf{S}_1$.  
This can be formulated as a conditional generation problem:
\begin{equation}
\mathbf{S}_2 = f_\theta(\mathbf{S}_1, \mathbf{C}, \mathbf{M}),
\end{equation}
where $f_\theta$ is the relighting model parameterized by $\theta$.

\noindent\textbf{Flow Matching.} 
To efficiently model complex illumination transformations in relighting, we build upon \textit{Wan2.1}~\cite{wan2025}, a foundation video generation model based on flow matching~\cite{lipman2022flow, esser2024scaling}. 
Flow matching formulates generative modeling as learning velocity fields between noise $x_0 \sim \mathcal{N}(0, I)$ and data $x_1$, using a linear interpolation:
\begin{equation}
x_t = t \cdot x_1 + (1 - t) \cdot x_0,\quad 
v_t = \frac{\mathrm{d}x_t}{\mathrm{d}t} = x_1 - x_0,
\label{flow_matching}
\end{equation}
where $t \in [0, 1]$ is sampled from a logit-normal distribution. The model learns to predict $v_t$ from $x_t$, conditioned on timestep $t$ and context $c$ (e.g., text embeddings), by minimizing the mean squared error:
\begin{equation}
\mathcal{L}_0 = \mathbb{E}_{x_0, x_1, t, c} \left\| v_\theta(x_t, t, c) - v_t \right\|_2^2.
\end{equation}

\noindent\textbf{Path Consistency Learning.} 
To further accelerate inference, we adopt path consistency learning~\cite{frans2024one}, which encourages consistent velocity predictions under larger integration steps. 
Given a velocity field $v_\theta$ and step size $d > 0$, we recursively define:
\begin{equation}
x_{t+d} = x_t + d \cdot v_\theta(x_t, t, d),
\end{equation}
Moreover, enforce two-step consistency using:
\begin{equation}
\mathcal{L}_{\text{fast}} = \mathbb{E}_{x_t, t, d} \left\| v_\theta(x_t, t, 2d) - \frac{1}{2} \left[ v_\theta(x_t, t, d) + v_\theta(x_{t+d}, t+d, d) \right] \right\|_2^2.
\label{fast_learning}
\end{equation}
This objective enables the model to learn shortcut-consistent velocity fields without separate teacher-student stages, allowing for fast and high-quality generation with arbitrary step budgets.

\begin{figure*}[t]
\centering
\includegraphics[width=\textwidth]{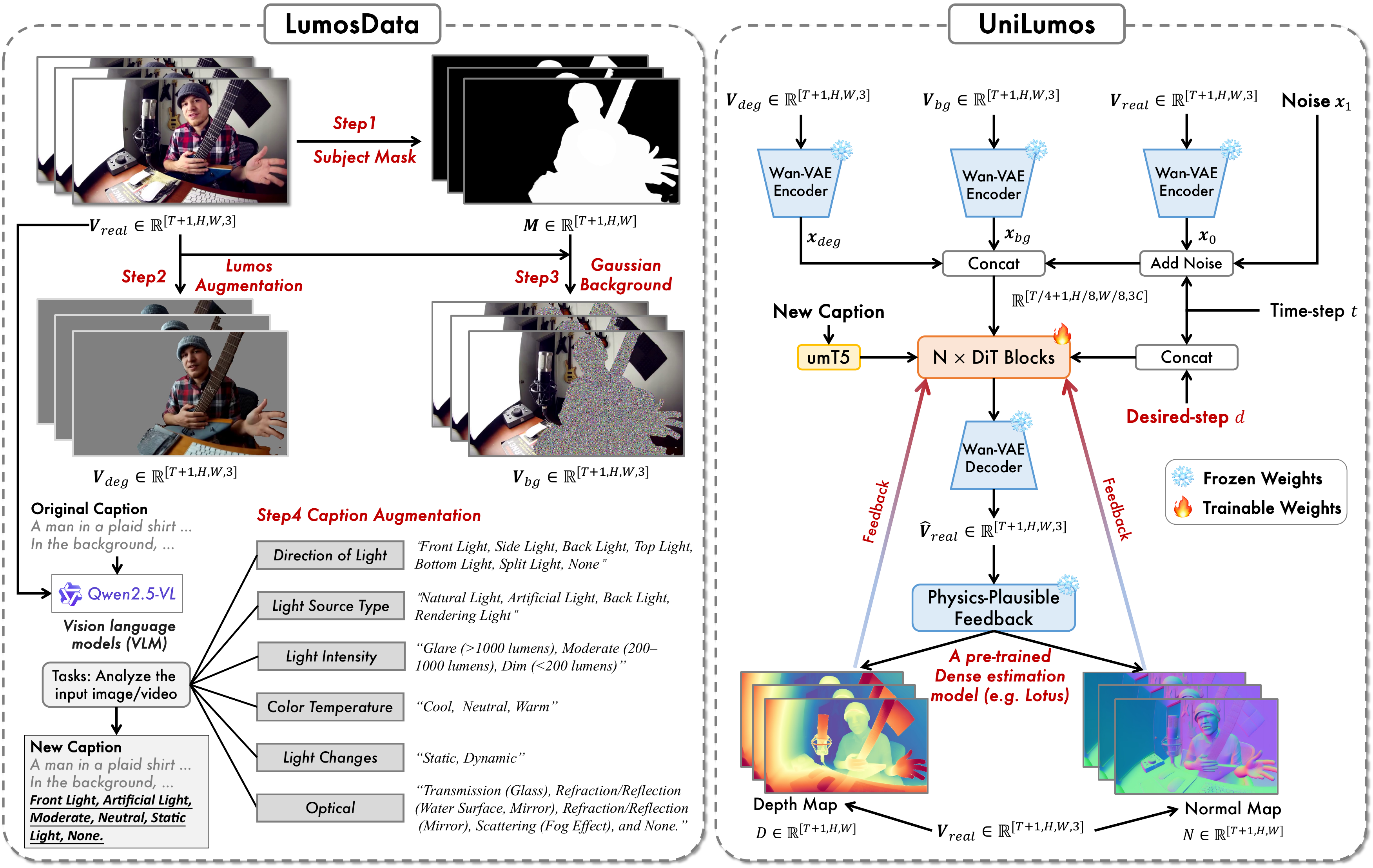}
\vspace{-4mm}
\caption{The overall pipeline of \textbf{UniLumos}. The left is \textit{LumosData}, our proposed data construction pipeline, which consists of four stages for generating diverse relighting pairs from real-world sources. The right shows the architecture of UniLumos, a unified framework for image and video relighting, designed to achieve physically plausible illumination control.}
\label{pipline}
\vspace{-4mm}
\end{figure*}
\section{Methodology}
We present \textbf{UniLumos}, a unified framework for physically plausible image and video relighting, as illustrated in Fig.~\ref{pipline}. 
Built upon Wan 2.1~\cite{wan2025}, a flow-matching diffusion model for video generation, UniLumos relights images and videos under user-specified lighting conditions—including reference images, video clips, or text prompts—while preserving scene content and temporal coherence.

To bridge the gap between semantic generation and physical correctness, UniLumos incorporates two key innovations: (1) a physics-plausible feedback that supervises the model with geometry signals from RGB space, and (2) a structured illumination annotation protocol that enables fine-grained control and evaluation. We jointly train the model with geometry-aware supervision and lighting-conditioned objectives, achieving high-quality and efficient few-step inference.

\subsection{Physics-Plausible Feedback}
While most relighting methods rely on photometric reconstruction or latent-space consistency, such signals offer limited geometric grounding—often resulting in misaligned shadows, implausible shading, and incorrect light directions. To address this, UniLumos enforces consistency between generated illumination and underlying scene geometry, promoting more realistic light–scene interactions.

As illustrated in Fig.~\ref{pipline} (right), we introduce a physics-plausible feedback that guides the generation process using geometry-aware supervision. This component complements the flow-matching architecture with explicit structural priors, enhancing physical plausibility without altering the model’s inference inputs. 
We adopt depth and surface normals as our supervisory targets due to their generality, accessibility, and strong disentanglement from illumination. Unlike shadow masks or material properties, which are often ambiguous, entangled with lighting, or costly to obtain, monocular depth and normal maps capture intrinsic scene structure. They can be reliably estimated by a pre-trained dense estimator (e.g., Lotus~\cite{he2024lotus}).

Specifically, after decoding the predicted latent variable into RGB frames via the \textit{Wan-VAE Decoder}~\cite{wan2025}, we extract estimated depth $\hat{\mathbf{D}} \in \mathbb{R}^{T \times H \times W}$ and normals $\hat{\mathbf{N}} \in \mathbb{R}^{T \times H \times W}$ using a frozen dense estimator. 
These are compared against pseudo-ground-truth maps $(\mathbf{D}, \mathbf{N})$ from the reference input to compute the geometry-aware feedback loss:
\begin{equation}
\mathcal{L}_{\text{phy}} = \mathbb{E}_{x_0, x_1, t, c} \left[
\mathbf{M} \odot 
\left(
\frac{\| \hat{\mathbf{D}} - \mathbf{D} \|_2}{\| \mathbf{D} \|_2} + 
\frac{\| \hat{\mathbf{N}} - \mathbf{N} \|_2}{\| \mathbf{N} \|_2}
\right) \right],
\end{equation}
where $\mathbf{M} \in \mathbb{R}^{T \times H \times W}$ denotes the foreground subject mask. This feedback encourages the model to align its lighting predictions with consistent structural interpretation while keeping inference lightweight and geometry-free.

However, the proposed physics-plausible feedback requires supervision in the RGB domain, which relies on high-quality predictions that are typically only available after full-step denoising has been completed. This poses a major computational bottleneck for standard diffusion models. To mitigate this, we adopt path consistency learning~\cite{frans2024one}, which reformulates denoising as a velocity regression task, thereby supporting practical training under a few-step regimes. Enforcing consistency between intermediate outputs and final predictions enables reliable geometric feedback without sacrificing inference efficiency. 

\subsection{Structured Illumination Annotation and Evaluation Benchmark}
In the problem formulation, the relighting task involves conditioning on a target illumination descriptor $\mathbf{C}$. 
However, most prior work treats $\mathbf{C}$ as unstructured prompts—such as text, images, or reference frames—offering limited control or interpretability. Moreover, conventional evaluation metrics, such as FID or LPIPS, focus on perceptual similarity but fail to capture lighting-specific discrepancies, such as shadow misalignment or intensity mismatches.

To address this, we construct LumosData, a scalable dataset pipeline that enriches $\mathbf{C}$ with structured lighting semantics. As shown in Fig.~\ref{pipline} (left), we extract relighting pairs from real-world videos. 
Given an input sequence $\mathbf{V}_{\text{real}} \in \mathbb{R}^{[T+1, H, W, 3]}$, we first obtain subject masks $\mathbf{M} \in \{0,1\}^{[T+1, H, W]}$ using BiRefNet~\cite{zheng2024birefnet} to isolate the foreground. We then apply a pre-trained relighting model such as IC-Light~\cite{zhang2025scaling} to generate synthetic relit versions $\mathbf{V}_{\text{deg}}$ under diverse lighting conditions, guided by a curated prompt set. To avoid entanglement with background semantics, we inpaint the background using Gaussian noise, ensuring clean illumination signals without introducing artifacts.

Beyond this relighting pipeline, LumosData introduces a structured six-dimensional annotation protocol that covers direction, intensity, color temperature, light source type, temporal dynamics, and optical phenomena. 

These attributes are automatically generated using vision-language models (e.g., Qwen2.5-VL~\cite{Qwen2.5-VL}) with carefully designed prompts, and are integrated into $\mathbf{C}$ to provide an enriched semantic label. 
This protocol serves dual purposes: 
(1) Fine-grained conditioning: During training, the model is guided by explicit lighting attributes embedded in $\mathbf{C}$, promoting more interpretable and controllable generation across scenarios.
(2) Attribute-aligned Benchmark: Building upon the same attribute protocol, we construct LumosBench. This automatic benchmark uses vision-language models to assess whether generated outputs accurately reflect intended lighting conditions, enabling interpretable, attribute-level controllability evaluation beyond pixel-based metrics.

Each training tuple is structured as $(\mathbf{V}_{\text{deg}}, \mathbf{V}_{\text{bg}}, \mathbf{M}, \mathbf{C}) \rightarrow \mathbf{V}_{\text{real}}$, where the model learns to restore realistic lighting consistent with the semantic cue $\mathbf{C}$ and structural context. LumosData introduces rich diversity in content and illumination by leveraging a variety of real-world sources. Built on Panda70M~\cite{chen2024panda}, we curate $\sim$110K high-quality video pairs and augment training with 1.2M additional relit images using IC-Light. This combination supports robust learning of physically plausible relighting without relying on expensive hardware or manual annotations. See more details in Appendix~\ref{app_dataset}.

\subsection{Joint Learning Objective and Training Strategy}
\textbf{Model Implementation.} 
The inputs of aligned videos $(\mathbf{V}_{\text{deg}}, \mathbf{V}_{\text{bg}}, \mathbf{V}_{\text{real}})$ are passed through a \textit{Wan-VAE Encoder}~\cite{wan2025} to obtain semantic latent representations $(\mathbf{x}_{\text{deg}}, \mathbf{x}_{\text{bg}}, \mathbf{x}_0)$. 
During training, we generate the noisy latent input $\mathbf{x}_t$ via Eq.~\ref{flow_matching}, and concatenate it with the strong conditional signals $\mathbf{x}_{\text{deg}}$ and $\mathbf{x}_{\text{bg}}$ along the channel dimension. This combined tensor is injected into the DiT blocks of the Wan backbone. 
Additionally, to support path-consistency learning, the diffusion step $t$ and the expected denoising step $d$ are appended as temporal condition vectors. 
All new projection and fusion layers are initialized with zero weights to preserve compatibility with the pre-trained Wan initialization and ensure stable optimization from the outset.

\textbf{Joint Objective.} Our training objective integrates three complementary losses to balance appearance fidelity, geometric consistency, and fast inference. The full loss is defined as:
\begin{equation}
\mathcal{L} = \lambda_0 \mathcal{L}_0 + \lambda_1 \mathcal{L}_{\text{fast}} + \lambda_2 \mathcal{L}_{\text{phy}},
\label{eq:total_loss_condensed}
\end{equation}
where $\mathcal{L}_0$ is the standard flow-matching loss that aligns the predicted velocity field with the ground-truth field, $\mathcal{L}_{\text{fast}}$ is the path consistency loss that improves model performance under few-step denoising regimes, and $\mathcal{L}_{\text{phy}}$ is a physics-guided loss that supervises the RGB outputs using estimated depth and normal maps. 
We adopt fixed weights of $\lambda_0 = 1.0$ and $\lambda_1 = \lambda_2 = 0.1$ for all experiments. This unified objective encourages the model to generate relit results that are photorealistic, temporally smooth, and physically grounded while supporting efficient inference without sacrificing output quality.



\begin{wrapfigure}{r}{0.5\textwidth}
\vspace{-8mm}
\begin{minipage}{0.48\textwidth}
\begin{algorithm}[H]
\caption{Loss Sampling Strategy per Iteration}
\label{alg:loss-schedule}
\begin{algorithmic}[1]
\REQUIRE Batch size $B$, total training samples
\FOR{each training iteration}
    \STATE Randomly sample 20\% of batch $\rightarrow$ $\mathcal{L}_{\text{fast}}$
    \STATE Compute path consistency loss:
    \STATE \hspace{0.5em} 3× forward, 1× backward
    \STATE Remaining 80\% $\rightarrow$ $\mathcal{L}_{0}$
    \STATE \quad Among those, 50\% $\rightarrow$ RGB reconstruction
    \STATE \quad Compute physics-guided loss $\mathcal{L}_{\text{phy}}$
\ENDFOR
\end{algorithmic}
\end{algorithm}
\end{minipage}
\end{wrapfigure}

\textbf{Training Strategy.} To balance physical supervision and training efficiency, we adopt a selective optimization strategy inspired by path consistency scheduling~\cite{frans2024one}. In each training iteration, we divide the batch based on supervision type, following an 80/20 split to avoid prohibitive costs from full supervision while still maintaining effective learning signals. As shown in Alg.~\ref{alg:loss-schedule}, 20\% of each batch is allocated to compute the path consistency loss $\mathcal{L}_{\text{fast}}$, which involves three forward passes and one backward pass to enforce consistency across timesteps. The remaining 80\% is used for the standard flow-matching loss $\mathcal{L}_0$, with 50\% of these samples further supervised using RGB-space geometry feedback via $\mathcal{L}_{\text{phy}}$ (i.e., depth and normal alignment). This probabilistic scheduling ensures high training throughput while allowing the model to benefit from multi-level supervision. To further enhance illumination diversity during training, we apply randomized lighting augmentations on the degraded subject $\mathbf{V}_{\text{deg}}$, which introduces realistic lighting variability without the need for explicitly paired captures.

\section{Experiments}
\subsection{Experimental Details}
\noindent\textbf{Training Details.} We adopt the Wan2.1-T2V-1.3B-480P \cite{wan2025} as the base model, and initialize all new trainable parameters with zeros to minimize its influence at the beginning of training. We use the AdamW optimizer with the learning rate of 1e-5 for training the entire framework. All the models are trained with a batch size of 8 for 5,000 iterations on 8 NVIDIA H20 GPUs (with 96GB RAM).

\noindent\textbf{Baselines.} We compare UniLumos against a range of representative image and video relighting methods. For image-based relighting, we include SwitchLight \cite{kim2024switchlight}, DiLightNet \cite{zeng2024dilightnet}, IC-Light \cite{zhang2025scaling}, and SynthLight \cite{chaturvedi2025synthlight}, which leverage various forms of latent modeling or light disentanglement to relight single images. 
For video relighting, we apply IC-Light via frame-by-frame and include Light-A-VideoCogVideoX-2B\cite{yang2024cogvideox} and another using Wan 2.1 T2V-1.3B~\cite{wan2025}. These baselines together represent state-of-the-art performance across both image and video relighting settings.

\noindent\textbf{Dataset.} For testing, we selected samples from the internal dataset, processed using the method described in Sec.~\ref{app_dataset}. These samples were evenly split: half for image generation at 768x512 resolution, and half for video generation at 480p resolution (832x480), with each video sample containing 49 frames. 
To further demonstrate the model’s generalization to non-human scenes, we conducted additional evaluations on two public object-centric relighting benchmarks: StanfordOrb \cite{kuang2023stanford} and Navi \cite{jampani2023navi}, which include objects and sculptures under a variety of lighting environments and are completely disjoint from our training data, and see more results in Appendix~\ref{add_main_results}.

\noindent\textbf{Evaluation metrics.} 
We evaluate relighting performance across three key dimensions: (1) visual fidelity: We assess image quality using Peak Signal-to-Noise Ratio (PSNR), Structural Similarity Index (SSIM), and Learned Perceptual Image Patch Similarity (LPIPS). For video relighting, we report the average metric across all frames. (2) temporal consistency: Following VBench~\cite{huang2024vbench}, we adopt the R-Motion metric, which measures temporal smoothness using motion priors from a pre-trained video frame interpolation model~\cite{li2023amt}. This captures the coherence of lighting transitions across frames. (3) lumos consistency: (i) Lumos Score, computed by applying the same caption-based lighting annotation protocol as in our LumosData construction. Six lighting attributes are predicted and compared with targets, and each is weighted equally to yield an average consistency score. (ii) Dense L2 Error, which quantifies the relative L2 error between predicted and reference depth/normal maps, estimated via a pre-trained geometry model (e.g., Lotus~\cite{he2024lotus}). This provides a physically grounded measure of illumination-geometry alignment.

\begin{table*}[t]
\centering
\caption{Quantitative comparison. \textbf{Bold} number indicate the best performance.}
\label{main_results}
\resizebox{\textwidth}{!}{
\begin{tabular}{l|ccc|c|cc}
\toprule
\multirow{2}{*}{\textbf{Model}} & 
\multicolumn{3}{c|}{\textbf{(a) Quality}} & \textbf{(b) Temporal Consistency} & \multicolumn{2}{c}{\textbf{(c) Lumos Consistency}}\\
& PSNR $\uparrow$ & SSIM $\uparrow$ & LPIPS $\downarrow$ 
& R-Motion$\downarrow$ & Avg. Score $\uparrow$ & Dense L2 Error $\downarrow$ \\
\midrule
\multicolumn{7}{c}{\textbf{Image Relighting}}\\
\midrule
SwitchLight \cite{kim2024switchlight}      
&20.483 &0.901 &0.094&-  &0.717&0.388\\
DiLightNet  \cite{zeng2024dilightnet}      
&21.894 &0.860 &0.131 &-  &0.682&0.401\\
IC-Light    \cite{zhang2025scaling}        
&24.316 &0.884 &0.108 &-  &0.703&0.447\\
SynthLight \cite{chaturvedi2025synthlight} 
&25.572 &0.905 &0.102 &-  &0.791&0.214\\
\midrule
\rowcolor{blue!10}  
\textbf{UniLumos} 
&\textbf{26.719} & \textbf{0.913} & \textbf{0.089} &- &\textbf{0.912} &\textbf{0.103}\\
\midrule
\multicolumn{7}{c}{\textbf{Video Relighting}}\\
\midrule
IC-Light Per Frame  \cite{zhang2025scaling} 
&20.132 &0.851 &0.133 &2.437 &0.672&0.432\\
Light-A-Video \cite{zhou2025light} + CogVideoX\cite{yang2024cogvideox}
&19.851 &0.859 &0.124 &1.784 &0.641&0.383\\
Light-A-Video \cite{zhou2025light}  + Wan2.1\cite{wan2025}
&20.784 &0.876 &0.129 &1.582 &0.682&0.371\\
\midrule
\rowcolor{blue!10}  
\textbf{UniLumos}                   
&\textbf{25.031} &\textbf{0.891} &\textbf{0.109} &\textbf{1.436} &\textbf{0.871} &\textbf{0.147}\\
\bottomrule
\end{tabular}
}
\end{table*}
\begin{figure*}[]
\centering
\includegraphics[width=\textwidth]{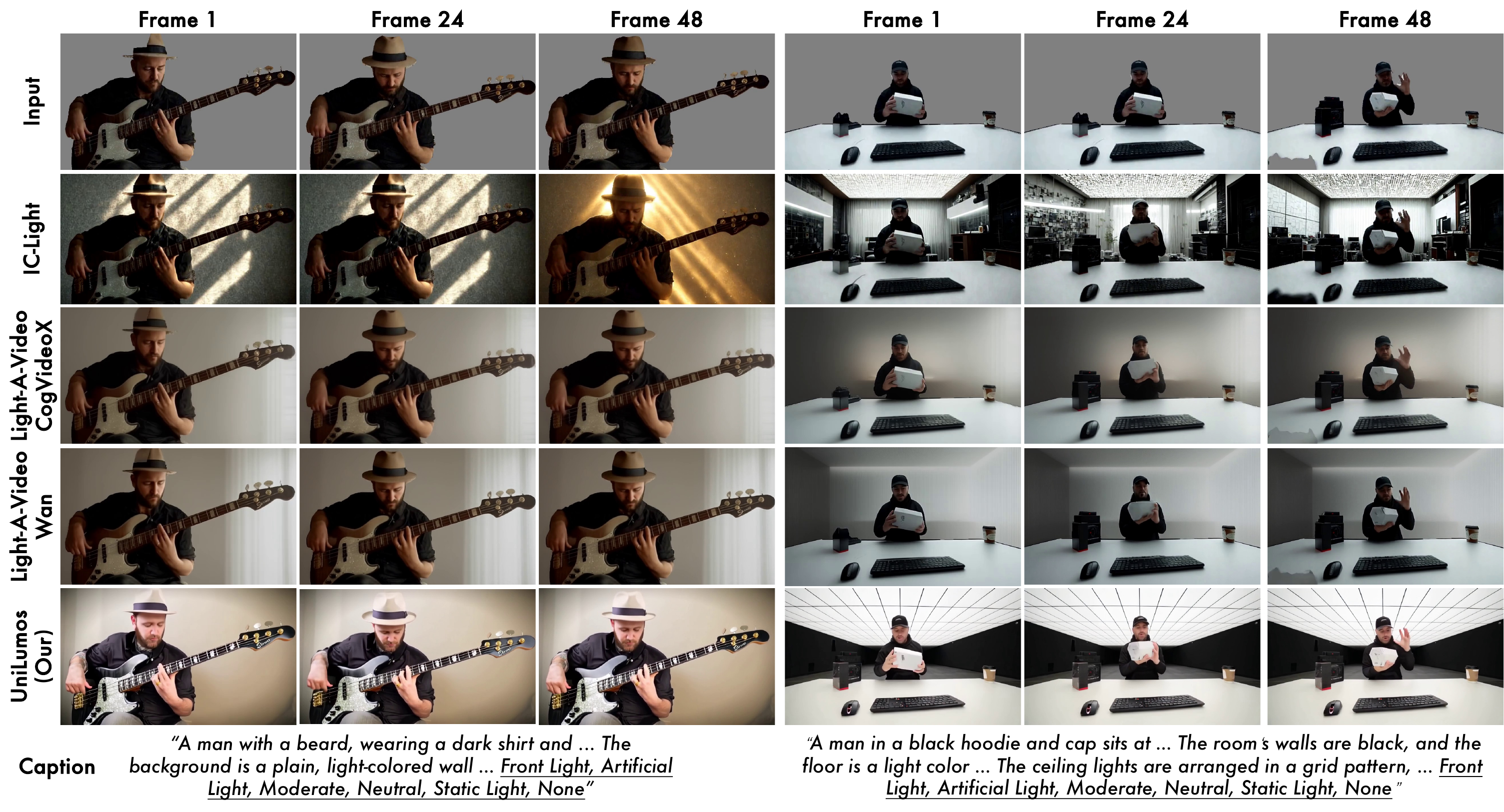}
\vspace{-6mm}
\caption{Qualitative comparison of baseline methods. Each method takes a subject video and a textual illumination description as input, generating the related subject with the corresponding background under the specified lighting condition.} 
\label{vis_main_results}
\vspace{-2mm}
\end{figure*}

\begin{figure*}[t]
\centering
\includegraphics[width=\textwidth]{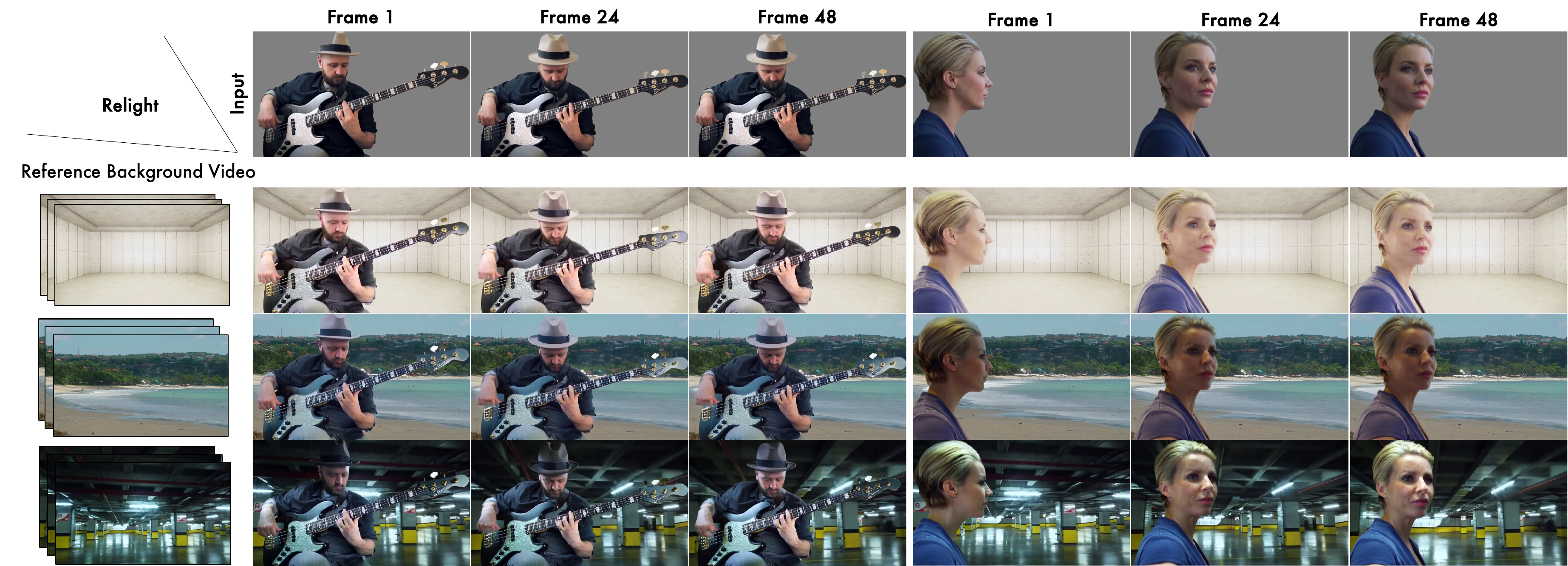}
\vspace{-6mm}
\caption{UniLumos performs physically plausible video relighting conditioned on different reference videos.} 
\label{vis_main_results_new}
\vspace{-2mm}
\end{figure*}

\subsection{Main Results}

\noindent\textbf{Quantitative Evaluation.}
 As shown in Tab.~\ref{main_results}, UniLumos delivers consistent improvements across three key dimensions: visual fidelity, temporal consistency, and physically grounded lighting alignment.
(1) Visual Fidelity.
 UniLumos produces higher-quality relighting results across both images and videos. Benefiting from structured lighting supervision and geometry-guided feedback, our model generates outputs with clearer shading, sharper details, and more coherent illumination compared to prior works.
(2) Temporal Consistency.
 For video relighting, UniLumos ensures smoother transitions and reduced flickering artifacts. Our use of flow-matching architecture and path consistency learning helps maintain stable lighting across frames, addressing a key limitation in frame-wise or training-free methods.

\begin{wrapfigure}{r}{0.5\textwidth}
\vspace{-8mm}
\centering
\includegraphics[width=0.45\textwidth]{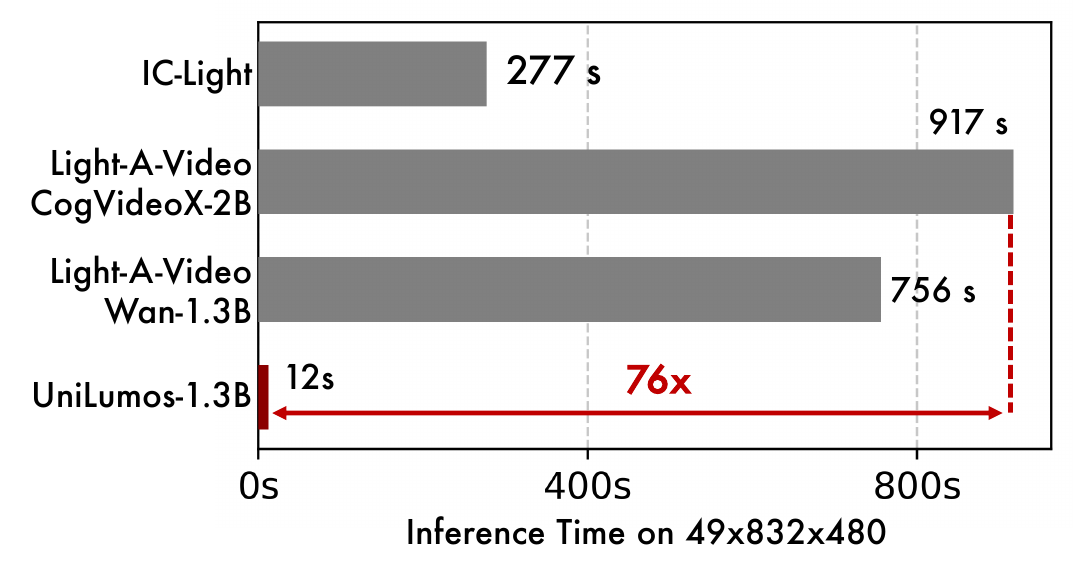}
\vspace{-5mm}
\caption{Comparison of inference time costs of different methods under the same settings.}
\label{fast_result}
\vspace{-4mm}
\end{wrapfigure}

(3) Lumos Consistency. Going beyond appearance-based evaluation, UniLumos aligns well with intended lighting semantics. Through structured caption conditioning and physics-guided training, the model better preserves lighting direction, tone, and geometry—validated by both vision-language alignment and dense geometric error metrics.

\noindent\textbf{Efficiency.} To assess inference efficiency, we evaluate the video relighting task under a standardized setting, generating 49-frame videos at a 480p resolution. As shown in Fig.~\ref{fast_result}, UniLumos achieves a significant speedup compared to prior methods, benefiting from its geometry-free inference and few-step generation. While existing models, such as Light-A-Video or IC-Light, require either iterative frame-by-frame processing or complex sampling schedules, UniLumos completes generation over 20 times faster without sacrificing visual fidelity or physical plausibility. This efficiency advantage makes it well-suited for real-time relighting applications and scalable deployment scenarios.

\noindent\textbf{Qualitative Results} We present qualitative comparisons in Fig. \ref {vis_main_results} and Fig. \ref{vis_main_results_new}, highlighting the advantages of UniLumos in terms of lighting realism, temporal coherence, and controllability. 
(1) Lighting Quality and Controllability: In Fig.~\ref{vis_main_results}, UniLumos produces lighting effects that better match the target description, capturing nuanced directional shadows, color tone, and intensity. Competing methods either fail to reflect the intended lighting change or produce overly uniform results that lack realism.
(2) Temporal Consistency: Compared to baseline methods such as frame-wise IC-Light and Light-A-Video, UniLumos achieves smoother frame transitions without flickering or structural distortion. This benefit arises from the joint modeling of space and time, which is further reinforced by physics-aware supervision and path consistency training.
(3) Foreground Detail Preservation: UniLumos preserves fine subject details—such as facial structure and clothing texture—better than baselines. For instance, Light-A-Video occasionally introduces deformation or identity drift, while our model maintains high fidelity over long sequences.
(4) Relighting with Reference Videos: Fig.~\ref{vis_main_results_new} showcases UniLumos conditioned on different reference videos. The model successfully adapts both global lighting direction and subtle spatial variations across scenes, demonstrating strong generalization under diverse illumination cues.

\noindent\textbf{LumosBench} To evaluate the fine-grained controllability of lighting generation, we introduce LumosBench, a structured benchmark that targets six core illumination attributes defined in our annotation protocol. Unlike prior works that treat lighting holistically or implicitly, LumosBench provides a disentangled, attribute-level evaluation, enabling precise diagnosis of model behavior under controllable lighting conditions. See more results in Appendix~\ref{add_benchmark}.
\begin{table*}[t]
\centering
\caption{Quantitative comparison. \textbf{Bold} number indicate the best performance.}
\label{main_results_ab}
\resizebox{\textwidth}{!}{
\begin{tabular}{c|ccc|c|cc}
\toprule
\multirow{2}{*}{\textbf{Model}} & 
\multicolumn{3}{c|}{\textbf{(a) Quality}} & \textbf{(b) Temporal Consistency} & \multicolumn{2}{c}{\textbf{(c) Lumos Consistency}}\\
& PSNR $\uparrow$ & SSIM $\uparrow$ & LPIPS $\downarrow$ 
& R-Motion$\downarrow$ & Avg. Score $\uparrow$ & Dense L2 Error $\downarrow$ \\
\midrule
\multicolumn{7}{c}{\textbf{Ablative Study}}\\
\midrule
w/o Depth Feedback   &23.472 &0.883 &0.118   &1.443 &0.870 &0.265\\
w/o Normal Feedback  &22.115 &0.874 &0.123   &1.446 &0.863 &0.173\\
w/o All Feedback     &21.433 &0.862 &0.139   &1.473 &0.859 &0.297\\
w/o Path Consistency &\textbf{25.317} &\textbf{0.902} &0.113 &1.438 &\textbf{0.875}&0.153\\
\midrule
\multicolumn{7}{c}{\textbf{Effect of Training Domain}}\\
\midrule
Only Video          &22.487&0.863&0.119&1.487&0.857&0.173\\
Only Image          &24.471&0.872&0.123&2.429&0.841&0.182\\
\midrule
\rowcolor{blue!10}  
\textbf{UniLumos} &25.031 &0.891 &\textbf{0.109} &\textbf{1.436}&0.871&\textbf{0.147}\\
\bottomrule
\end{tabular}
}
\end{table*}
\begin{figure*}[t]
\centering
\includegraphics[width=\textwidth]{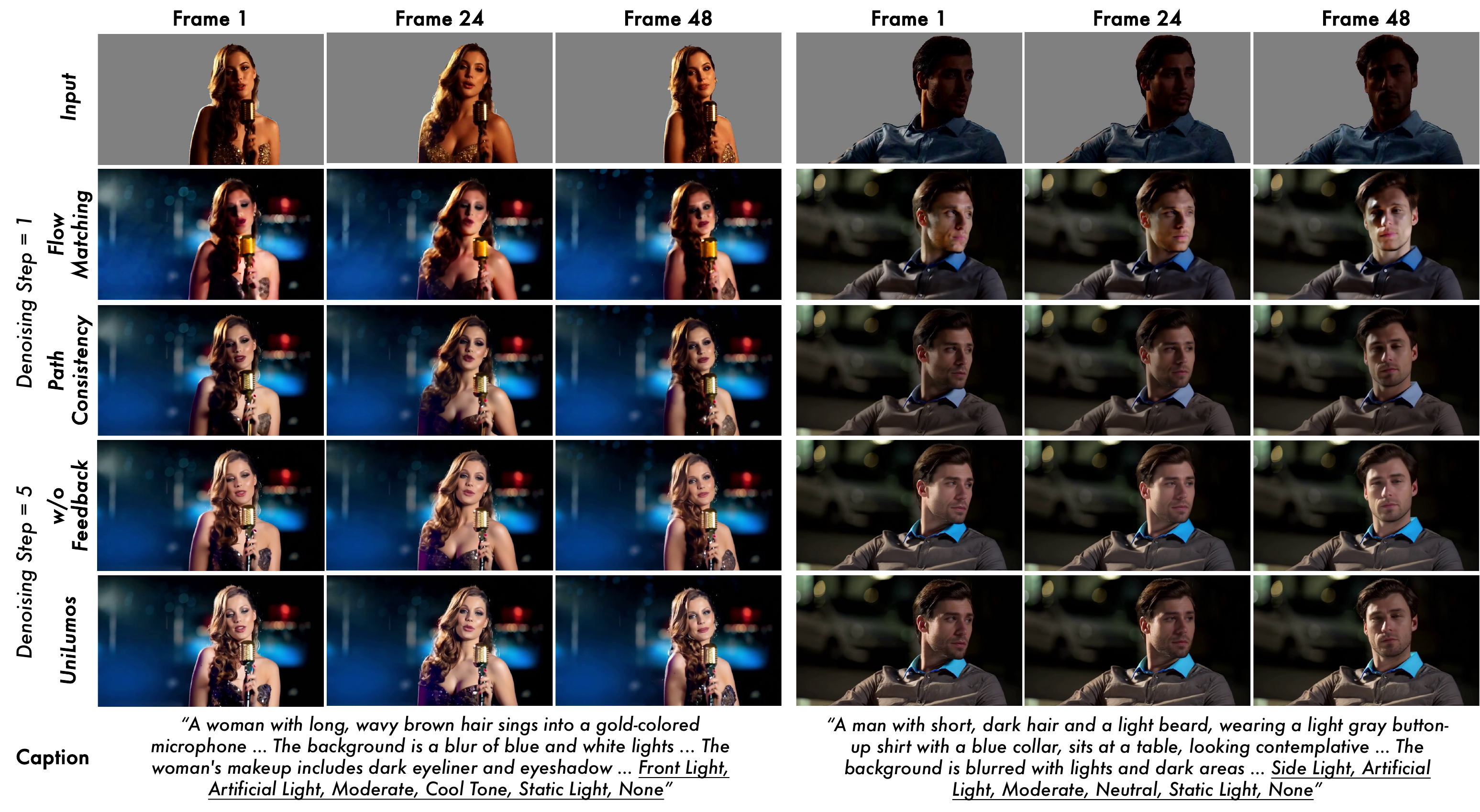}
\vspace{-6mm}
\caption{Ablation study. We compare the effects of different components under few-step denoising. For 1-step, we show the impact of flow-matching with and without path consistency. For 5-step, we visualize results before and after introducing physics-plausible feedback.}
\label{vis_ablation_study}
\end{figure*}
\subsection{Ablation Study}

As shown in Tab.~\ref{main_results_ab} and Fig.~\ref{vis_ablation_study}, we conduct ablation studies to analyze the effectiveness of different components. \textbf{Physics-Guided Feedback.} Removing both depth and normal feedback (w/o All Feedback) leads to significant degradation in both image quality and physical consistency, confirming the necessity of our physics-guided loss. Notably, omitting only normal supervision causes a larger drop than removing depth, suggesting that surface orientation plays a more critical role than distance in shaping light–shadow interactions. \textbf{Path Consistency Learning.} Excluding this component (w/o Path Consistency) yields only minor drops in physical metrics while maintaining competitive SSIM and LPIPS scores. This shows that path consistency incurs little performance cost but offers substantial efficiency benefits in few-step regimes, justifying its inclusion. \textbf{Training Modality.} To evaluate the effectiveness of our unified training paradigm, we compare domain-specific variants: training solely on videos leads to poor visual quality, while image-only training sacrifices temporal smoothness. In contrast, our unified approach strikes the best balance—achieving high-quality and temporally coherent relighting across both input types. 

\section{Conclusion}
We introduce UniLumos, a unified framework for physically plausible image and video relighting. It aligns illumination with scene geometry via RGB-space depth and normal supervision, improving shadow accuracy and spatial consistency. To enhance controllability and evaluation, we propose a structured six-dimensional lighting annotation protocol, enabling fine-grained conditioning and physically grounded assessment through VLMs. Experiments show that UniLumos achieves superior relighting quality, physical consistency, and inference efficiency.

\newpage
\section*{Acknowledgment}
This work was supported by Damo Academy through Damo Academy Research Intern Program.
\bibliographystyle{plain}
\bibliography{mybib}

\newpage
\appendix
\begin{center}
    {\Large \textbf{Appendix of UniLumos}}
\end{center}

In this appendix, we provide additional details to complement the main paper.  
First, we explain the motivation behind introducing physics-plausible feedback in \textbf{Sec.~\ref{app_why}}.  
Next, we present the detailed pipeline of our proposed LumosData relighting data construction process in \textbf{Sec.~\ref{app_dataset}}.  
Then, \textbf{Sec.~\ref{add_results}} contains three additional experimental results on the public dataset (Sec.~\ref{add_main_results}) and the LumosBench (Sec.~\ref{add_benchmark}).
\textbf{Sec.~\ref{app_vis}} provides additional qualitative results to further illustrate the effectiveness of UniLumos.
Finally, \textbf{Sec.~\ref{limitation}} discusses the limitations of our method.

\section{Physics-Plausible Feedback}\label{app_why}
To further clarify the motivation and design behind our physics-plausible feedback mechanism, we present a breakdown of key questions and considerations addressed during its development.

\textbf{Q1: What is the motivation for introducing physical constraints in relighting?} 

\textbf{A1:} The primary goal of relighting is to generate visually plausible illumination under new lighting conditions. However, many diffusion-based methods lack explicit physical modeling, leading to artifacts such as overexposed highlights, misaligned shadows, or inconsistent light directions. Introducing physical constraints serves as a refinement mechanism that aligns generated light with the scene’s underlying geometry. This helps enforce realism and spatial consistency in illumination, which is especially critical under complex lighting or HDR scenarios.

\textbf{Q2: Why are depth and normal maps chosen as the targets for physical supervision?} 

\textbf{A2:} Depth and surface normals are among the most accessible and general-purpose dense scene attributes. By design, these estimations intentionally suppress fine-scale lighting effects to focus on intrinsic geometry. This makes them ideal for supervising relighting, where the goal is to decouple geometry from illumination and enforce spatial structure in lighting behavior. In the proposed UniLumos, we align the predicted lighting with reference geometry by minimizing the L2 norm error between generated and reference-aligned depth/normal maps via a pre-trained dense estimation model (e.g., Lotus \cite{he2024lotus}) with frozen parameters. This provides a simple yet effective metric to quantify physical plausibility.

\textbf{Q3: Why are alternative physical signals—such as albedo, shadow, or material—not used instead?} 

\textbf{A3:} While albedo, shadow masks, and material properties can provide rich supervision, they come with significant drawbacks. Albedo and shadow estimation often rely on inverse rendering and suffer from domain sensitivity or ambiguity. Material annotations are expensive and dataset-dependent. Moreover, many of these properties are entangled with illumination, making them less reliable as supervisory signals. In contrast, depth and normals can be predicted from monocular images with high availability and generalize well across scenes, offering a favorable balance between supervision quality and computational cost.

\textbf{Q4: Why are depth and normal maps used as training-time constraints rather than as model inputs?} 

\textbf{A4:} While it is possible to condition the model directly on estimated depth and normal maps, doing so increases the input dimensionality and model complexity. It would also introduce a dependency on external estimators during inference, complicating the pipeline and potentially propagating errors. Instead, we use them as supervision signals during training. This design keeps the inference pipeline simple—relying only on image and lighting condition inputs—while still allowing the model to learn geometry-aware behaviors. The supervision acts as a form of inductive bias, guiding the model toward physically plausible outputs without requiring additional input channels at the test phase.

\section{Details of Datasets}\label{app_dataset}

\noindent\textbf{Step 1: Subject Mask.} Given an input video $\mathbf{V}_{\text{real}}\in\mathcal{R}^{[T+1,H,W,3]}$, we first extract per-frame subject masks $\mathbf{M}\in\mathcal{R}^{[T+1,H,W]}$ using BiRefNet~\cite{zheng2024birefnet}. These subject masks allow us to isolate the target subject foreground and the target background. 

\noindent\textbf{Step 2: Lumos Augmentation.}  
To simulate diverse lighting degradations for training, we relight each subject sequence under multiple lighting conditions using a pre-trained 2D relighting model, such as IC-Light~\cite{zhang2025scaling}. This operation is applied independently to each frame of the subject region, resulting in a degenerated video $\mathbf{V}_{\text{deg}} \in \mathcal{R}^{[T+1,H,W,3]}$. To generate rich illumination variations, we refer to the description of light and shadow given by IC-Light~\cite{zhang2025scaling}, as listed in Tab.~\ref{tab:quick-prompts}, which serve as the semantic guidance for image-level relighting and the light source directions. For each input video, we randomly sample 5 prompts and 3 directions, forming $5 \times 3 = 15$ unique prompt-direction pairs. The relighting is applied only to the subject region, extracted using the subject masks $\mathbf{M} \in \mathcal{R}^{[T+1,H,W]}$ from \textbf{Step 1}. Notably, we randomly sample one degradation condition from the 15 prompt-direction combinations for each subject in each iteration. This strategy reduces training cost while exposing the model to diverse illumination patterns, thereby improving generalization.
    
\begin{table*}[t]
\centering
\caption{Lighting-related textual prompts used in Lumos Augmentation from IC-Light~\cite{zhang2025scaling}. Each prompt can be combined with different canonical light directions during training.}
\label{tab:quick-prompts}
\begin{tabular}{c|l|l}
\toprule
\textbf{ID} & \textbf{Lighting Prompt} & \textbf{Example Light Direction} \\
\midrule
1  & sunshine from window                      & None \\
2  & neon light, city                          & Left Light \\
3  & sunset over sea                           & Right Light \\
4  & golden time                               & Top Light \\
5  & sci-fi RGB glowing, cyberpunk             & Bottom Light \\
6  & natural lighting                          &  \\
7  & warm atmosphere, at home, bedroom         &  \\
8  & magic lit                                 &  \\
9  & evil, gothic, Yharnam                     &  \\
10 & light and shadow                          &  \\
11 & shadow from window                        &  \\
12 & soft studio lighting                      &  \\
13 & home atmosphere, cozy bedroom illumination&  \\
14 & neon, Wong Kar-wai, warm                  &  \\
\bottomrule
\end{tabular}
\end{table*}

\noindent\textbf{Step 3: Gaussian Background.}  To provide external lighting context during training, we generate a background video $\mathbf{V}_{\text{bg}} \in \mathbb{R}^{[T+1, H, W, 3]}$ to accompany the relit subject. Instead of relying on complex inpainting-based synthesis (e.g., ProPainter~\cite{zhou2023propainter}, DiffuEraser~\cite{li2025diffueraser}), we adopt a simple yet effective strategy by filling the background with either pure color or Gaussian noise. This design avoids injecting semantic or structural priors, allowing the model to focus solely on illumination learning.

Specifically, for each frame $t \in [1, T+1]$ and channel $c \in \{R, G, B\}$, we first define the background region using the subject mask $\mathbf{M}_t \in \mathbb{R}^{H \times W}$ obtained in \textbf{Step 1}. Let $\Omega_{bg}^t = \{(i,j) \mid \mathbf{M}_t(i,j) = 0\}$ denote the set of background pixels. We compute the mean and standard deviation of background pixel intensities as:
\begin{equation}
\left\{
\begin{aligned}
& \mu_c^t = \frac{1}{|\Omega_{bg}^t|} \sum_{(i,j) \in \Omega_{bg}^t} \mathbf{V}_t(i,j,c),\\
& \sigma_c^t = \sqrt{ \frac{1}{|\Omega_{bg}^t|} \sum_{(i,j) \in \Omega_{bg}^t} \left( \mathbf{V}_t(i,j,c) - \mu_c^t \right)^2 }
\end{aligned}
\right.
\end{equation}

\begin{figure*}[t]
\centering
\includegraphics[width=\textwidth]{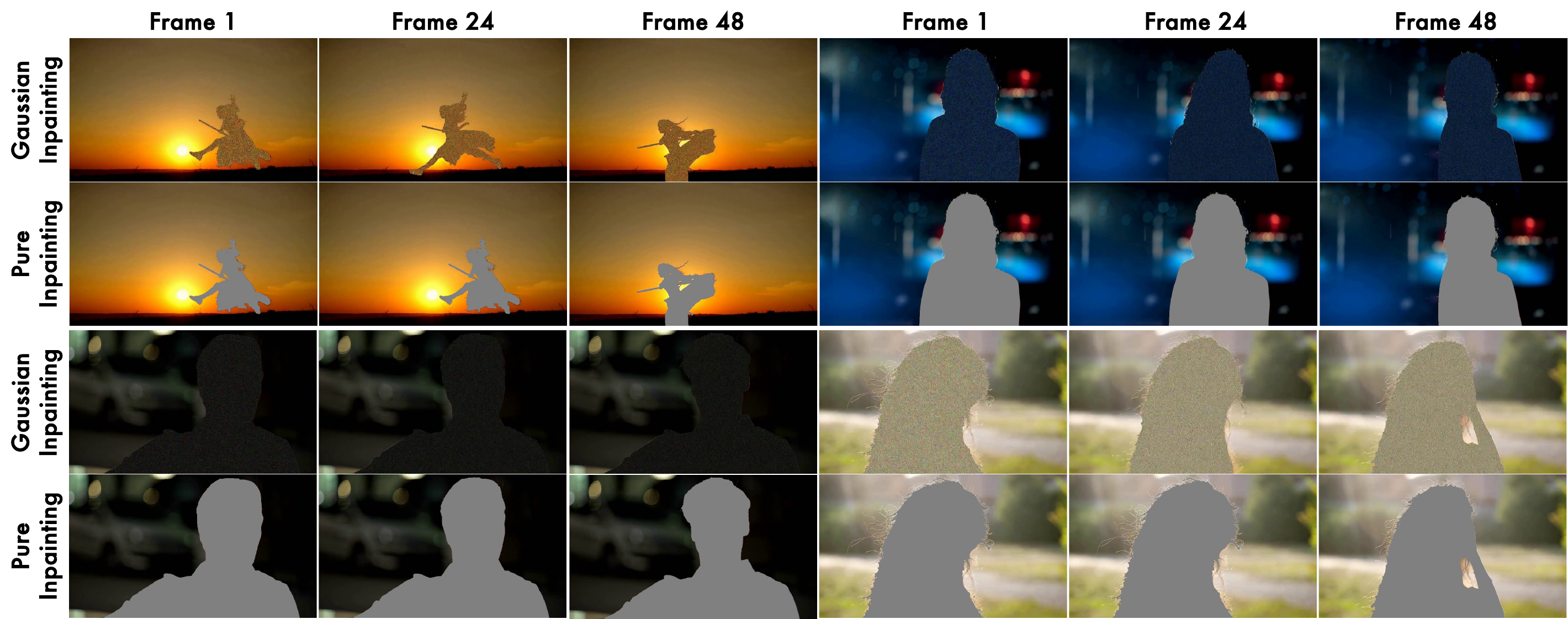}
\vspace{-6mm}
\caption{Comparison of background inpainting strategies of four representative cases. Here, \textit{Gaussian Inpainting} fills the background using random noise sampled with the same mean and variance as the subject region, ensuring statistical consistency. \textit{Pure Inpainting} directly fills the background with a uniform color (i.e., gray), without modeling spatial or color variation. The Gaussian strategy provides more realistic signal distribution and accelerates early-stage convergence in training.}
\label{fig:bg_vis}
\end{figure*}

We then fill the background with pixel-wise samples from a Gaussian distribution:
\begin{equation}
\mathbf{V}_{bg}^t(i,j,c) \sim \mathcal{N}(\mu_c^t, (\sigma_c^t)^2), \quad \forall (i,j) \in \Omega_{bg}^t.
\end{equation}
This procedure ensures that the background region maintains a similar color distribution to the original video while avoiding structural detail that may bias learning. 
As shown in Fig. \ref{fig:bg_vis}, for comparison, we also test a variant that uses pure-color background, where each background pixel is set to $\mu_c^t$ (i.e., $\sigma_c^t = 0$). In practice, we observe that such statistically consistent placeholders—particularly Gaussian-filled ones—accelerate early-stage convergence during training. We attribute this to the reduced visual complexity and improved normalization behavior, which make the model less sensitive to background variation. The resulting $\mathbf{V}_{bg}$ serves as a clean, distribution-aligned conditioning signal for the relighting network.

\noindent\textbf{Step 4: Caption Augmentation.}  
In addition to relighting augmentation, we generate lighting-aware captions to provide rich semantic supervision aligned with physical lighting behavior. Specifically, we leverage Qwen2.5-VL~\cite{Qwen2.5-VL}, a vision-language model with fine-grained visual reasoning capabilities, to analyze each input video and generate structured captions describing its lighting attributes. The input to Qwen2.5-VL consists of the original video and its corresponding scene-level caption. We then apply a custom-designed prompt (see Listing~\ref{lst:prompt}) to steer the model toward predicting six categories of lighting-related labels as shown in Tab.~\ref{tab:light-attributes}, including all subcategories and their physical interpretations. The output of this process is a structured caption $\mathbf{C}$ for each video, formatted as a dictionary mapping the six categories to their predicted labels (see example in Listing~\ref{lst:prompt}). These structured captions serve as auxiliary supervision and evaluation labels in later stages, helping the model better align with interpretable physical lighting semantics. They also enhance downstream controllability and facilitate attribute-based retrieval or evaluation.

\begin{table*}[h]
\centering
\caption{Classification criteria and definitions for light-related scene attributes.}
\label{tab:light-attributes}
\resizebox{\textwidth}{!}{
\begin{tabular}{c|>{\columncolor{palegreen}}c|>{\columncolor{palecream}}c}
\toprule
\textbf{Primary Category} & \textbf{Subcategory} & \textbf{Definition} \\
\midrule
\multirow{7}{*}{Direction of Light} 
  & Front Light & The light source is positioned directly in front of the subject, illuminating it head-on. \\  
  & Side Light & The light source is positioned at a 90-degree or 45-degree angle to the subject, coming from the side. \\
  & Back Light & The light source is located behind the subject, directed towards the camera. \\
  & Top Light & The light source is positioned directly above the subject, casting light downwards. \\
  & Bottom Light & The light source is positioned below the subject, casting light upwards. \\
  & Split Light & The light source illuminates one side of the subject while leaving the other side in shadow. \\
  & Ambient Light Without Clear Direction & Ambient light is non-directional, uniformly illuminating the environment from multiple sources. \\
\midrule
\multirow{3}{*}{Light Source Type}
  & Natural Light & Illumination from nature without human intervention, varying with time of day, weather, and location. \\
  & Artificial Light & Human-made light sources (e.g., bulbs, LEDs) used in indoor/outdoor spaces for functional or artistic effects. \\
  & Rendering Light & Digitally simulated light in CGI, games, or animations using techniques like ray tracing. \\
\midrule
\multirow{3}{*}{Light Intensity}
  & Glare & Extremely bright light over 1000 lumens that can cause discomfort or obscure detail. \\
  & Moderate & Balanced lighting (200–1000 lumens), suitable for most activities and comfortable viewing. \\
  & Dim & Low lighting under 200 lumens, often cozy but may reduce visibility and detail recognition. \\
\midrule
\multirow{3}{*}{Color Temperature}
  & Cool Tone & 5000K–10000K; bluish hues, common in daylight or overcast scenes. \\
  & Neutral & 4000K–5000K; balanced light with no strong blue or yellow tint. \\
  & Warm Tone & 2000K–4000K; reddish or yellowish hues, typical in sunrise/sunset or indoor lighting. \\
\midrule
\multirow{3}{*}{Light Changes in Time}
  & Static Light & Illumination remains constant in both intensity and direction over time. \\
  & Dynamic Light (Intensity Changing) & Light intensity changes gradually over time (e.g., dawn to daylight). \\
  & Dynamic Light (Moving Source) & Direction of light changes due to movement of light source (e.g., headlights, stage lights). \\
\midrule
\multirow{4}{*}{Optical Phenomena}
  & Transmission (Glass) & Light passes through transparent materials like glass, with possible scattering or absorption. \\
  & Refraction/Reflection (Water Surface, Mirror) & Light bends or reflects at water or mirror surfaces, altering its direction. \\
  & Scattering (Fog Effect) & Light diffuses through particles like fog or mist, reducing visibility. \\
  & None & No significant optical phenomena are observed in the scene. \\
\bottomrule
\end{tabular}
}
\end{table*}

\newpage
\begin{lstlisting}[language=Python, caption={Prompt Definition}, label={lst:prompt}]
SYSTEM_PROMPT = """
You are a helpful, respectful and honest assistant. Always answer as helpfully as possible, while being safe. Your answers should not include any harmful, unethical, racist, sexist, toxic, dangerous, or illegal content. Please ensure that your responses are socially unbiased and positive in nature.\n\nIf a question does not make any sense, or is not factually coherent, explain why instead of answering something not correct. If you don't know the answer to a question, please don't share false information.
"""

PROMPT = """
Role: You are an expert in image/video light and shadow analysis, good at analyzing light and shadow from multiple angles.
...

Tasks: Analyze the input image/video, provide corresponding classification results for the following multiple categories, and return them in the specified output format.

1. Direction of Light: 
    Task 1: Analyze the image and classify the light source direction as front light, side light, back light, top light, bottom light, or split light. Identify the angle of the light source relative to the subject, and describe its effect on shadow formation.

2. Light Source Type: 
    Task 2: Analyze the image and classify the light source type as either Natural Light, Artificial Light, or Rendering Light.

3. Light Intensity:
    Task 3: Analyze the image/video to assess the light intensity present. Classify the light intensity into three categories: Glare, Moderate, and Dim. Special attention should be given to situations where bright light sources may create a glaring effect even in otherwise dim environments.

4. Color Temperature:
    Task 4: Analyze the image/video to assess the color temperature present. Classify the color temperature into three categories: Cool Tone, Neutral, and Warm Tone. 

5. Light Changes in Time: 
    Task 5: Analyze the video to assess light changes over time. Classify the light changes into two main categories: Static Light and Dynamic Light. For Dynamic Light, further categorize it into two subtypes: Intensity Gradient and Moving Light Source.

6. Optical Phenomena: 
    Task 6: Analyze the image/video with a focus on the specific scene to assess the optical phenomena present. Pay close attention to scenarios involving glass, water surfaces, mirrors, and fog. Classify the phenomena into the following categories: Transmission (Glass), Refraction/Reflection (Water Surface, Mirror), Refraction/Reflection (Mirror), Scattering (Fog Effect), and None.

Guidelines:

1. Accuracy: Assign each tag to the most appropriate category and subcategory.
2. Multiple Tags: If an action fits multiple categories, assign all relevant tags.
3. Comprehensiveness: Capture all detectable dynamic attributes without omissions.
4. JSON Validity: Ensure the output JSON is correctly formatted and adheres to the specified structure.

Example Output:
{
    "Direction of Light": "Front Light",
    "Light Source Type": "Artificial Light",
    "Light Intensity": "Moderate",
    "Color Temperature": "Cool Tone",
    "Light Changes in Time": "Dynamic Light (Intensity Changing Light)",
    "Optical Phenomena": "Transmission (Glass)"
}
"""
\end{lstlisting}

\section{Additional Experimental Results}\label{add_results}

\subsection{Additional Main Results}\label{add_main_results}

To further demonstrate the model’s generalization to non-human scenes, we conducted additional evaluations on two public object-centric relighting benchmarks: StanfordOrb \cite{kuang2023stanford} and Navi \cite{jampani2023navi}. These datasets include objects and sculptures under a variety of lighting environments and are completely disjoint from our training data. StanfordOrb contains canonical 3D scanned objects such as the Stanford bunny and dragon, while Navi includes a wide range of everyday objects like containers, toys, and mugs. As shown in Tab. \ref{tab:stanfordorb_performance} (StanfordOrb) and Tab. \ref{tab:navi_performance} (Navi), UniLumos achieves state-of-the-art results across perceptual (LPIPS), structural (SSIM), and physical (R-Motion) metrics, despite the significant domain gap and without any test-time fine-tuning, outperforming all baselines.
\begin{table}[h]
    \centering
    \caption{Quantitative comparison on the StanfordOrb dataset. \textbf{Bold} number indicate the best performance.}
    \label{tab:stanfordorb_performance}
    \begin{tabular}{lcccc}
        \toprule
        \textbf{Model} & \textbf{PSNR} $\uparrow$ & \textbf{SSIM} $\uparrow$ & \textbf{LPIPS} $\downarrow$ & \textbf{R-Motion} $\downarrow$ \\
        \midrule
        IC-Light Per Frame \cite{zhang2025scaling} & 24.132 & 0.914 & 0.126 & 1.742 \\
        Light-A-Video \cite{zhou2025light} + CogVideoX \cite{yang2024cogvideox}& 25.617 & 0.923 & 0.108 & 1.279\\
        Light-A-Video \cite{zhou2025light} + Wan2.1 \cite{wan2025}& 25.784 & 0.926 & 0.104 & 1.241\\
        \midrule
        \textbf{UniLumos} & \textbf{26.512} & \textbf{0.934} & \textbf{0.097} & \textbf{1.103}\\
        \bottomrule
    \end{tabular}
\end{table}

\begin{table}[h]
    \centering
    \caption{Quantitative comparison on the Navi dataset. \textbf{Bold} number indicate the best performance.}
    \label{tab:navi_performance}
    \begin{tabular}{lcccc}
        \toprule
        \textbf{Model} & \textbf{PSNR} $\uparrow$ & \textbf{SSIM} $\uparrow$ & \textbf{LPIPS} $\downarrow$ & \textbf{R-Motion} $\downarrow$ \\
        \midrule
        IC-Light Per Frame \cite{zhang2025scaling} & 22.021 & 0.883 & 0.125 & 1.974 \\
        Light-A-Video \cite{zhou2025light} + CogVideoX \cite{yang2024cogvideox} & 23.912 & 0.891 & 0.121 & 1.378 \\
        Light-A-Video \cite{zhou2025light} + Wan2.1 \cite{wan2025} & 23.474 & 0.903 & \textbf{0.116} & 1.341 \\
        \midrule
        \textbf{UniLumos} & \textbf{24.977} & \textbf{0.911} & 0.120 & \textbf{1.203}\\
        \bottomrule
    \end{tabular}
\end{table}

\subsection{LumosBench: An Attribute-level Controllability Benchmark}\label{add_benchmark}

To evaluate the fine-grained controllability of lighting generation, we introduce LumosBench, a structured benchmark that targets six core illumination attributes defined in our annotation protocol. Unlike prior works that treat lighting holistically or implicitly, LumosBench provides a disentangled, attribute-level evaluation, enabling precise diagnosis of model behavior under controllable lighting conditions.

Specifically, we construct a set of 2k test prompts, each consisting of a video and a structured caption designed to isolate one lighting attribute at a time, while holding other variables constant. These prompts span six categories—direction, light source type, intensity, color temperature, temporal dynamics, and optical phenomena—with multiple subtypes per category (e.g., front/side/back for direction). 
This design facilitates controlled and interpretable evaluation across lighting axes that are often conflated in prior datasets.

To assess alignment between intended and generated lighting attributes, we use the vision-language model Qwen2.5-VL~\cite{Qwen2.5-VL} to analyze relit outputs and classify whether the target attribute is correctly expressed. Each dimension is scored independently, and the final controllability score is the average across all six dimensions.

\begin{table*}[t]
\centering
\vspace{-2mm}
\caption{Quantitative comparison of the attribute-level controllability. \textbf{Bold} number indicate the best performance.}
\label{main_results_lumos_ab}
\resizebox{\textwidth}{!}{
\begin{tabular}{l|c|cccccc|c}
\toprule
\textbf{Model} 
&\#Params 
&Direction   
&Light Source Type 
& Intensity & Color Temperature & Temporal Dynamics & Optical Phenomena & Avg. Score \\ 
\midrule
\multicolumn{9}{c}{\textbf{General Models}}\\
\midrule
LTX-Video \cite{hacohen2024ltx}          &1.9B &0.794&0.644&0.487&0.708&0.487&0.403&0.587\\
CogVideoX \cite{yang2024cogvideox}       &5.6B &0.837&0.692&0.552&0.739&0.532&0.449&0.634\\
HunyuanVideo \cite{kong2024hunyuanvideo} &13B  &0.863&0.741&0.599&0.802&0.655&0.481&0.690\\
Wan2.1\cite{wan2025}                    &1.3B  &0.842&0.685&0.436&0.741&0.504&0.433&0.607\\
Wan2.1\cite{wan2025}                    &14B &0.871&0.794&0.674&\textbf{0.829}&\textbf{0.737}&0.505&0.735\\
\midrule
\multicolumn{9}{c}{\textbf{Specialized Models}}\\
\midrule
IC-Light Per-Frame  
\cite{zhang2025scaling} 
&0.9B &0.793&0.547&0.349&0.493&0.284&0.339&0.468\\
Light-A-Video \cite{zhou2025light} + CogVideoX\cite{yang2024cogvideox}
&2.9B &0.787&0.581&0.327&0.536&0.493 &0.373 &0.516\\
Light-A-Video \cite{zhou2025light} + Wan2.1\cite{wan2025}
&2.2B&0.801&0.603&0.361&0.582&0.557 &0.412 &0.553\\
\midrule
\textbf{UniLumos} w/o lumos captions 
&1.3B &0.868&0.774&0.529&0.798&0.543&0.457&0.662\\
\rowcolor{blue!10}
\textbf{UniLumos} 
&1.3B &\textbf{0.893}&\textbf{0.847}&\textbf{0.832}&0.813&0.662&\textbf{0.592}&\textbf{0.773}\\
\bottomrule
\end{tabular}
}
\vspace{-4mm}
\end{table*}


\textbf{General vs. Specialized Models.} 
Tab.~\ref{main_results_lumos_ab} presents results for both general-purpose and task-specific relighting models. This benchmark allows us not only to assess overall lighting quality but also to dissect a model's ability to interpret and respond to individual lighting controls. Among general models, Wan 14B shows the highest raw capability, demonstrating the strength of large-scale pretraining for visual generation. Notably, our fine-tuned Wan 1.3B variant achieves substantial gains across all six lighting dimensions, surpassing even much larger models. This highlights the benefit of relighting-specific supervision: fine-tuning on LumosData with structured lighting annotations significantly enhances the model's ability to reason about illumination in a controllable and disentangled manner.

By contrast, specialized relighting models consistently underperform despite being designed for lighting manipulation. This is primarily due to the limitations of their base architectures—typically smaller and trained from scratch or on narrow domains—resulting in weaker generalization to diverse lighting attributes. While they may encode some prior knowledge of lighting physics (e.g., through latent constraints), their restricted modeling capacity hinders semantic alignment with user-intended lighting conditions. These findings underscore the importance of starting from a strong pre-trained backbone and introducing structured, high-level lighting supervision to achieve controllable and physically plausible relighting.

\textbf{Effectiveness of Structured Captions.} 
Within the specialized group, we conduct an ablation to assess the importance of our proposed lumos captions. \textbf{w/o lumos captions} uses only vanilla scene-level captions during training, omitting structured lighting tags. The performance drop—particularly in controllable dimensions like intensity and optical phenomena—confirms that our semantic annotations play a key role in teaching the model fine-grained illumination control. Compared to strong baselines, UniLumos achieves superior scores across nearly all dimensions, demonstrating the impact of LumosBench in pushing model understanding and control of illumination.


\section{Additional Result Visualization}\label{app_vis}

\begin{figure}[t]
\centering
\includegraphics[width=0.7\textwidth]{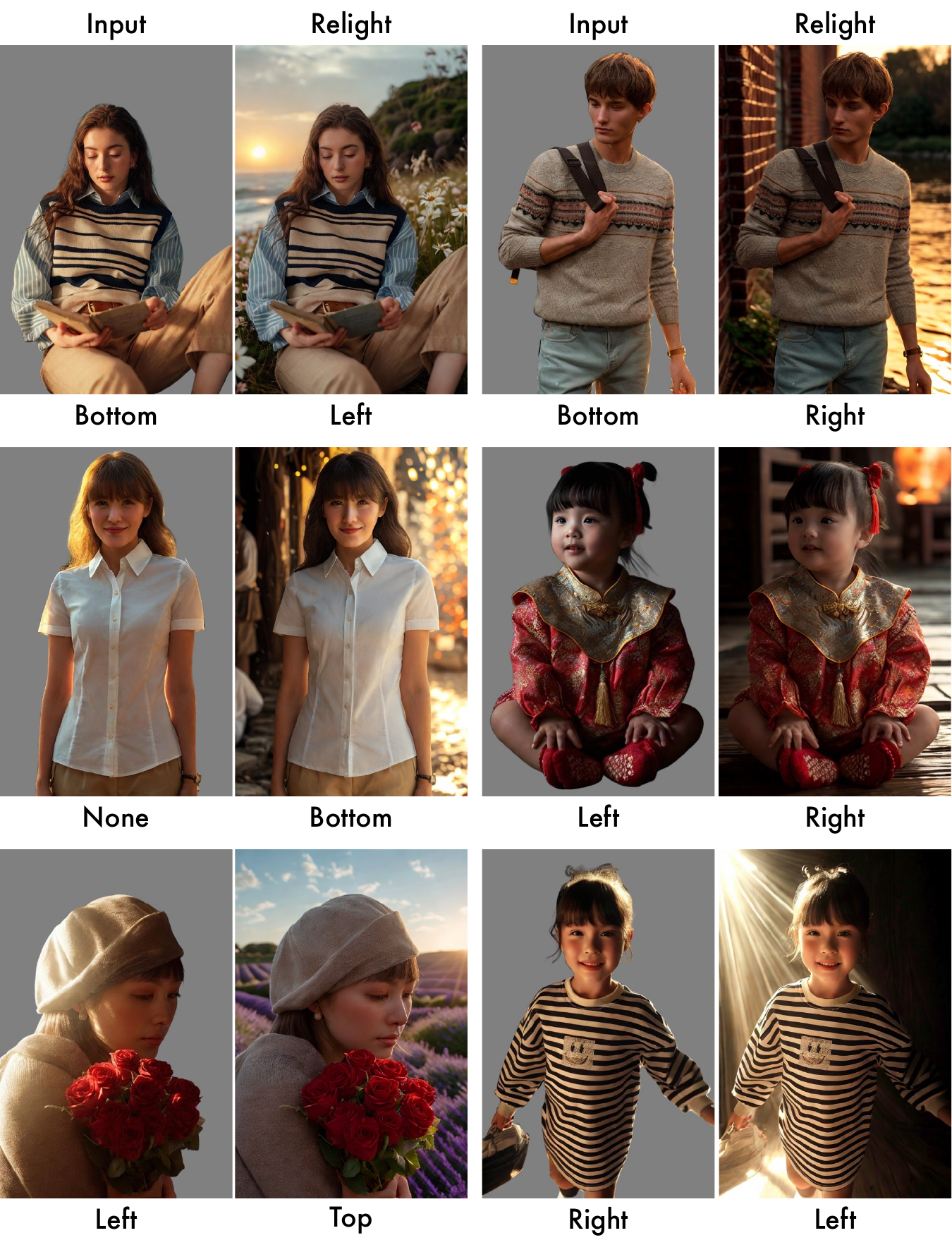}
\vspace{-4mm}
\caption{UniLumos performs physically plausible image relighting, conditioned on textual prompts.}
\label{result_image}
\end{figure}

We present additional image relighting results in Fig. \ref{result_image}.

\begin{figure}[t]
\centering
\includegraphics[width=\textwidth]{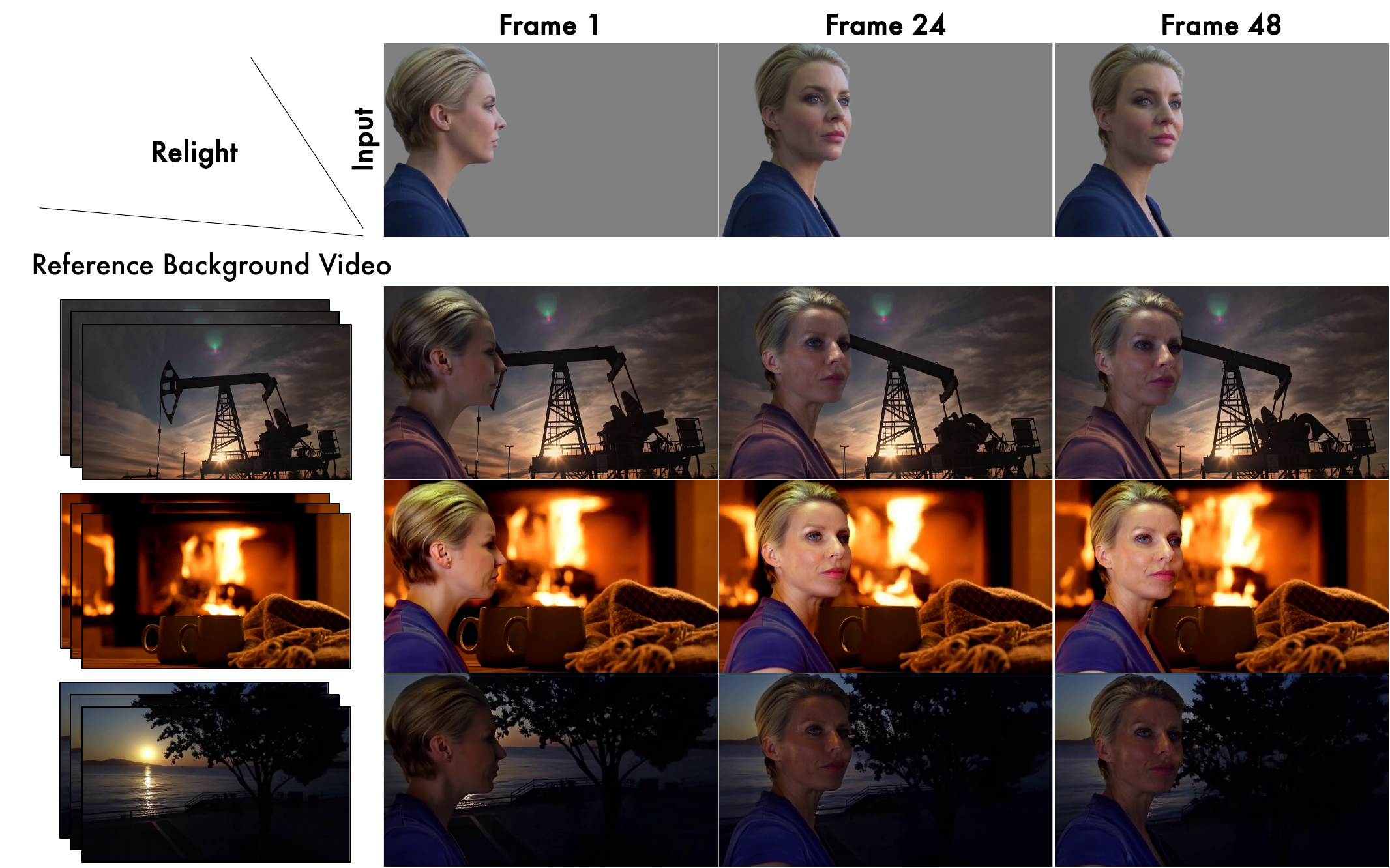}
\vspace{-4mm}
\caption{UniLumos performs physically plausible video relighting, conditioned on reference videos.}
\label{app_video_1}
\end{figure}

\begin{figure}[t]
\centering
\includegraphics[width=\textwidth]{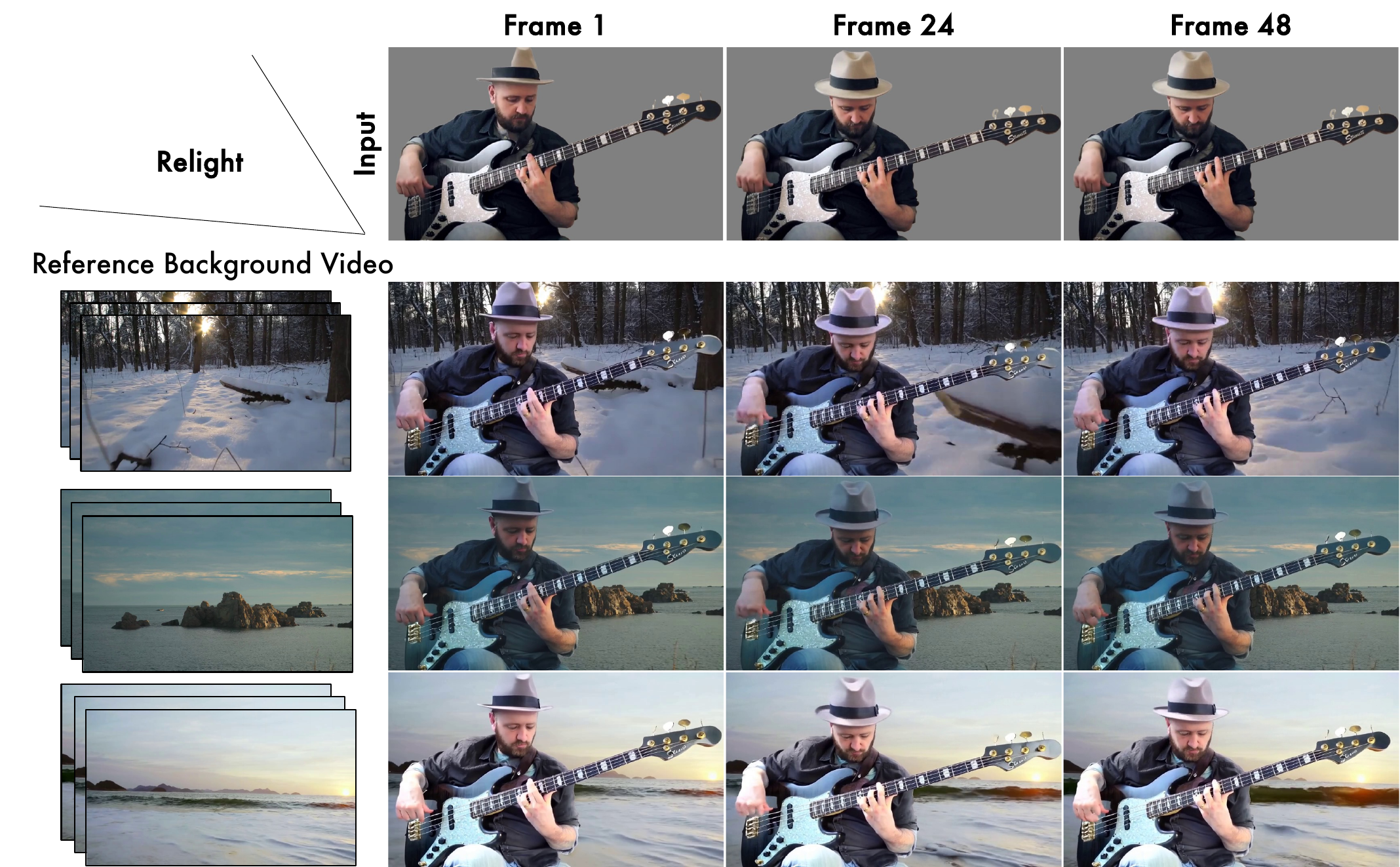}
\caption{UniLumos performs physically plausible video relighting, conditioned on reference videos.}
\label{app_video_2}
\end{figure}

We present additional background-conditioned video relighting results in Fig. \ref{app_video_1} and Fig. \ref{app_video_2}.

\section{Limitation and Future Work}\label{limitation}
UniLumos is still limited by a broader challenge—achieving physically precise and controllable relighting. UniLumos enforces geometry-aware consistency (e.g., shadows aligned with depth and normals) but does not yet produce physically quantifiable lighting outputs such as radiance or illuminance. Future work may explore finer control over lighting, including editable key lights, intensity ramps, and environmental reflections.


\end{document}